\documentclass{article}
\usepackage{ijcai18}

\usepackage{times}
\usepackage{soul}
\usepackage{url}
\usepackage[hidelinks]{hyperref}
\usepackage[utf8]{inputenc}
\usepackage[small]{caption}

\usepackage{graphicx}
\usepackage{amsmath}
\usepackage{subcaption}
\usepackage{amsfonts}
\usepackage{amsthm}

\usepackage{paralist}

\usepackage{tikz}
\usepackage{pgfplots}
\usepackage{framed}
\usepackage{tabu}

\pgfplotsset{compat=1.14}

\newtheorem{theorem}{Theorem}

\title{What Game Are We Playing?  \\ End-to-end Learning in Normal and Extensive Form Games}

\author{
Chun Kai Ling, 
Fei Fang, 
J. Zico Kolter 
\\ 
School of Computer Science, Carnegie Mellon University \\
chunkail@cs.cmu.edu,
feif@cs.cmu.edu,
zkolter@cs.cmu.edu
}

\begin{document}

\maketitle

\begin{abstract}
Although recent work in AI has made great progress in solving large, zero-sum, extensive-form games, the underlying assumption in most past work is that the parameters of the game itself are known to the agents.  This paper deals with the relatively under-explored but equally important ``inverse" setting, where the parameters of the underlying game are \emph{not} known to all agents, but must be learned through observations.  We propose a differentiable, end-to-end learning framework for addressing this task.  In particular, we consider a regularized version of the game, equivalent to a particular form of quantal response equilibrium, and develop 1) a primal-dual Newton method for finding such equilibrium points in both normal and extensive form games; and 2) a backpropagation method that lets us analytically compute gradients of all relevant game parameters through the solution itself.  This ultimately lets us learn the game by training in an end-to-end fashion, effectively by integrating a ``differentiable game solver" into the loop of larger deep network architectures.
We demonstrate the effectiveness of the learning method in several settings including poker and security game tasks.
\end{abstract}

\section{Introduction}
\label{sec:intro}

Recent work in artificial intelligence has led to huge advances in methods for solving large-scale, zero-sum, extensive form games, both from methodological and applied standpoints.  From the algorithmic approach, methods based on Counterfactual Regret Minimization \cite{zinkevich2008regret} and first-order methods for game solving \cite{kroer2017theoretical}, have enabled solutions to larger and larger games.  In terms of applications, there have been a number of recent breakthroughs, including exceeding human performance in no-limit poker \cite{brown2017superhuman,moravvcik2017deepstack}, essentially weakly solving limit poker \cite{bowling2015heads}, work in security games with applications to infrastructure security \cite{pita2009using}, and many others.
However, virtually all this progress in game theoretic approaches to large games has operated under the assumption that the parameters of the game are known to the solvers, and that the main challenge is simply finding the optimal strategy.  In contrast, in many real world scenarios, certain elements of the game (e.g., payoff matrices, chance node probabilities, etc), are unknown to some of the agents prior to the game.  For example, in security games, we may want to understand the underlying payoffs of an adversary, rather than just their observed strategy, to better understand how aspects of the game can be manipulated or changed to get a desirable outcome. 

\indent In this paper, we propose an end-to-end framework for learning the parameters of uncertain games (both for normal-form and extensive-form games), purely by observing the actions of the agents.  Although there has been a great deal of work at the intersection of game theory and reinforcement learning \cite{busoniu2008comprehensive,bowling2000analysis}, most game-theoretic analysis either assumes that the payoffs underlying the game are known (this is the standard game theory setting), or forgoes trying to learn an explicit and complete representation of the game and instead looks for merely learning agent strategies that will perform well \cite{letchford2009learning,vorobeychik2007learning,fearnley2015learning}.
However, in many cases when the true underlying payoffs of the agents are \emph{not} known, our primary goal is precisely to recover or understand the payoffs.
 The few exceptions that focus on learning the payoffs often rely on special structures of the game (e.g., symmetry in multiplayer setting \cite{vorobeychik2007learning}), or querying the best response of the agent with unknown payoffs by asking other agents to play carefully designed strategies \cite{blum2014learning,letchford2009learning}.
However, the general problem of learning game parameters by observing actions is still under-explored.  One of the most closely-related works to our own is the Computational Rationalization framework \cite{waugh2011computational}, though 1) our approach differs in how the utilities/payoffs are modeled; and 2) we crucially focus heavily on the extensive form settings, whereas this past work considered only normal form games.

The crux of our approach is to consider the \emph{quantal response equilibrium} (QRE), a generalization of Nash equilibrium (NE) that includes some possibility of agents acting suboptimally. We show that the solution of the QRE is a \emph{differentiable} function of the game payoff matrix, and backpropagation can be computed analytically via implicit differentiation.  We develop a solver that jointly solves the QRE for two-player zero-sum games using a primal-dual Newton Method, and allows us to compute the derivatives of agent actions with respect to the underlying payoff matrix.  This enables us to develop end-to-end learning approaches that can infer the payoff matrix or other parameters underlying a game merely from samples of the agents acting according to their QREs.  \footnote{Naturally, there are some questions here about when games are identifiable or not; in general, the answer is no, because e.g. many payoff matrices can lead to identical strategies or policies for the agent, so we do not expect to always be able to recover a true underlying payoff matrix. However, we show that for various classes of parametrized games, our approach \emph{is} able to recover the true underlying payoffs of different agents.}  More generally, the method allows for (both normal form and extensive form) game-solving to be integrated as a module in deep learning systems, a strategy that can find use in multiple application areas.  



We demonstrate the effectiveness of our approach on several domains: a toy normal-form game where payoffs depend on external context; a one-card poker game (with a small representation in strategic form, but which would already be too large to solved in normal form); and a security resource allocation game, which is an extensive-form generalization of defender-attacker game in security domain.  In all settings, we show that our approach is able to learn, solely from observed actions, the relevant underlying parameters of the game, such as the payoff matrices or (agent belief over) chance node probabilities.  We believe this represents a substantial step forward in understanding how game theoretic methods can be applied to uncertain settings, where the ``true'' parameters of the game are unknown to an agent. Due to space constraints, supplementary material and appendices are available at arXiv.\footnote{https://arxiv.org/abs/1805.02777} 


\section{Learning and Quantal Response in Normal Form Games}
\begin{figure}[t]
\centering
    \includegraphics[scale=0.35]{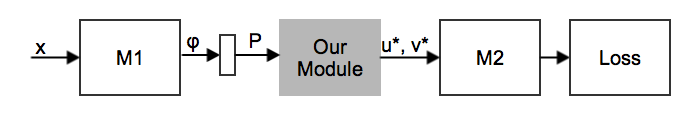}
    \caption{An example architecture utilizing our proposed module.}
   \label{fig:ex_architecture}
\end{figure}

Our game-solving module provides all the elements required to perform differentiable learning through the game solution. 
The resulting learning approach learns a mapping from \textit{contextual features} $x$ to payoff matrices $P$ and computes equilibrium strategies $(u^*, v^*)$ under a new set of contextual features. An example architecture is presented in Figure~\ref{fig:ex_architecture}. Here, $P$ is parameterized by a domain-dependent low-dimensional vector $\phi$, which is dependent on a differentiable function $M_1(x)$. Similarly, the loss function is taken after applying any differentiable $M_2(u^*, v^*)$. For the remainder of the paper, we focus on zero-sum games, which capture a wide class of adversarial environments.

We begin by considering normal form games. Although normal form games have limited real-world utility due to the fact that they can only handle relatively small-scale settings, the game solver and learning approach in this restricted setting captures much of the intuition and basic methodology of our approach.
\label{sec:formulation}
\subsection{Zero-Sum Normal Form Games}
\label{sec:learning_zsg}
In two-player zero-sum game with payoff matrix $P$, a classic min-max formulation to compute the NE is as follows
\begin{align}
    \min_{u} \max_{v}& \quad u^T P v \\
    \text{subject to} &  \quad 1^Tu=1, u \ge 0 \\
    & \quad  1^Tv=1, v \ge 0,
\end{align}
$u$ and $v$ denote the (mixed) strategies employed by the min and max player respectively. The solution $(u^*,v_0)$ to this optimization problem and the solution $(u_0,v^*)$ of the corresponding problem with inversed player order (i.e., $\min_v \max_u u^T P v$) forms the Nash equilibrium $(u^*,v^*)$.

Here we present an introduction to our approach considering the case where the payoff matrix $P$ is \emph{not} known a priori. $P$ could represent either a single fixed but unknown payoff matrix, or, in a more complex setting, depend on some external context $x$. For example, in anti-poaching games \cite{fang2016deploying}, $P$ depends on temperature and precipitation. In general, however, we consider the case where we observe samples of actions $a^{(i)}$, $i=1,\ldots,N$, consisting of observed actions, from one or both players, sampled from the equilibrium strategies $(u^*, v^*)$.  The goal is to recover the true underlying payoff matrix $P$, or a function form $P(x)$ depending on the current context.

\subsection{Quantal Response Equilibria}
\label{sec:QRE}
While extremely powerful both theoretically and as a modeling tool, the NE is poorly-suited for our purposes because:
\begin{enumerate}
    \item NEs are overly strict. In practice, many payoff matrices result in actions never being played. This tends to be overly restrictive and does not adequately describe real-world scenarios where players are boundedly rational.
    \item NEs in zero-sum games may not be unique. This leads to difficulties when resolving which NE to select.
    \item NEs are discontinuous with respect to $P$ -- a small change in $P$ can lead to jumps in $u^*, v^*$. This precludes integrating the technique into differentiable learning procedures.
\end{enumerate}

To address these issues, in our learning setting we propose to model the player's action with the quantal response equilibria \cite{mckelvey1995quantal} instead. In general, QRE models situations where payoff matrices are injected with some noise. Specifically, we consider the \emph{logit} equilibrium, where payoffs are perturbed by samples from a Gumbel distribution. The smoothness of QRE makes gradient-based approaches feasible \cite{amin2016gradient}. It is known that for zero-sum games, the logit equilibrium obeys the fixed point \footnote{In the QRE, there is an additional rationality parameter $\lambda$. In this work, we fix $\lambda=1$ throughout.}
\begin{equation}
\label{eqn:logit}
\begin{aligned}
u^*_i &= \frac{\exp(-Pv)_i}{\sum_{q \in [n]}\exp(-Pv)_q}, \quad v^*_j= \frac{\exp(P^Tu)_j}{\sum_{q \in [m]}\exp(P^Tu)_q}. 
\end{aligned}
\end{equation}
It is further known that for a fixed opponent strategy, the logit equilibrium corresponds to a strategy regularized by the Gibbs entropy~\cite{mertikopoulos2016learning}. Since the Gibbs entropy is strictly convex, the regularized best response is unique. 
\subsection{End-to-End Learning}
In order to integrate zero-sum game solvers into an end-to-end learning framework, we need a method for ``differentiating through'' the game solution itself; that is, we need to compute the Jacobian (or more precisely, compute the Jacobian-vector products needed for backpropagation) of the quantal equilibrium solution with respect to the payoff matrix.  Our method for doing so relies on techniques from differential calculus, and is a relatively straightforward extension of similar approaches to differentiating through optimization problems.  However, as a prelude to the more involved extensive form solution that we will discuss shortly, we describe our method in some detail, which involves both a particular approach to solving the QRE and to differentiating through its solution.
\subsubsection{QRE Solver}
We observe that finding the fixed point in \eqref{eqn:logit} is equivalent to solving the regularized min-max game
\begin{equation}
\begin{aligned}
    \min_{u \in \mathbb{R}^n} 
    \max_{v \in \mathbb{R}^m} & \quad
    u^T P v - H(v) + H(u) \\
    \text{subject to} & \quad 1^Tu = 1, \quad 1^Tv = 1,
\end{aligned}
\end{equation}
where $H(y)$ is the Gibbs entropy $\sum_i y_i \log y_i$. Notice that the non-negative constraints are implicit from the entropy term, and that the entropy regularization renders the equilibrium continuous with respect to $P$. Intuitively, entropy regularization encourages players to play more randomly, and no action has probability $0$. Furthermore, since the objective is strictly a convex-concave problem, it has a unique saddle point which corresponds to $(u^*, v^*)$. 

This formulation leads to a solver for the QRE for two-player zero-sum games, using a primal-dual Newton Method. 
To begin, the KKT conditions for the above problem are
\begin{equation}
\begin{aligned}
\label{eqn:KKT}
    Pv + \log(u) + 1 + \mu1 &= 0 \\
    P^Tu - \log(v) - 1 + \nu1 &= 0 \\
    1^T u = 1, \qquad 
    1^T v &= 1,
\end{aligned}
\end{equation}
where $\mu, \nu$ are Lagrange multipliers for the equality constraints on $u, v$ respectively. Following Newton's method, we get the following update rule, which provides a convergent method for computing the QRE for 2 player zero-sum games
\begin{align}
\label{eqn:newton}
    Q
    \begin{bmatrix}
    \Delta u \\
    \Delta v \\
    \Delta \mu \\
    \Delta \nu 
    \end{bmatrix} = 
    -\begin{bmatrix}
    Pv + \log u + 1 + \mu 1 \\
    P^Tu - \log v - 1 + \nu 1 \\
    1^Tu - 1 \\
    1^v - 1 
    \end{bmatrix},
    \intertext{where $Q$ is the Hessian of the Lagrangian, given by}
    Q = \begin{bmatrix}
    \text{diag}(\frac{1}{u}) & P & 1 & 0 \\
    P^T & -\text{diag}(\frac{1}{v}) & 0 & 1 \\
    1^T & 0 & 0 & 0 \\
    0 & 1^T & 0 & 0 
    \end{bmatrix}.
\end{align}
\subsubsection{Differentiating Through QRE Solutions}
The QRE solver also provides a method for computing the necessary Jacobian-vector products.  The derivation follows in a similar manner to recent work in differentiating equality-constrained optimization problems \cite{gould2016differentiating,johnson2016composing,amos2017optnet} (the only difference being the min-max objective instead of a pure minimization objective, but since we compute differentials via the KKT conditions, the differences are minor).  Specifically, given the solution $(u^*,v^*)$ to the QRE, and considering some loss function $L(u^*, v^*)$ (for example, the log-likelihood of some observed data given this equilibrium probabilities), we show here how to compute the gradient of the loss with respect to the payoff $P$.
In particular, taking differentials of the KKT conditions and rearranging leads to the following expression
\begin{align}
    Q
    \begin{bmatrix}
    \mathsf{d}u \enskip
    \mathsf{d}v \enskip
    \mathsf{d}{\mu} \enskip
    \mathsf{d}{\nu} \enskip
    \end{bmatrix}^T &=
    \begin{bmatrix}
    -\mathsf{d}Pv \enskip  
    -\mathsf{d}P^Tu \enskip 
    0 \enskip
    0 
    \end{bmatrix}^T.
\end{align}
For small changes denoted by $\mathsf{d}u, \mathsf{d}v$, we have 
\begin{equation*}
\begin{aligned}
\mathsf{d}L &= 
\begin{bmatrix}
\nabla_u L & \nabla_v L & 0 & 0
\end{bmatrix}
\begin{bmatrix}
\mathsf{d}u \enskip  \mathsf{d}v \enskip  \mathsf{d}{\mu} \enskip \mathsf{d}{\nu}
\end{bmatrix}^T \\
&= 
\begin{bmatrix}
\nabla_u L & \nabla_v L & 0 & 0
\end{bmatrix}
Q^{-1}
\begin{bmatrix}
    -\mathsf{d}Pv \enskip  
    -\mathsf{d}P^Tu \enskip 
    0 \enskip
    0 
    \end{bmatrix}^T \\
&= 
%
\begin{bmatrix}
    v^T\mathsf{d}P^T \enskip 
    u^T \mathsf{d}P \enskip 
    0 \enskip
    0 
\end{bmatrix}
Q^{-1}
\begin{bmatrix}
-\nabla_u L \enskip -\nabla_v L \enskip 0 \enskip 0
\end{bmatrix}^T,
\end{aligned}
\end{equation*}
where the last step is from symmetry  of $Q$. This expression governs how small changes in $\mathsf{d}P$ affect $L$. For example, we may obtain the change in $L$ after perturbing a single entry in $P$. Applying this procedure to all entries in $P$, simplifying and taking limits as $\mathsf{d}P$ is small yields
\begin{align}
\label{eqn:gradL_P}
    \nabla_P L = y_u v^T + u y_v^T,
\end{align}
where 
\begin{align*}
\begin{bmatrix}
    y_u \enskip
    y_v \enskip
    y_{\mu} \enskip
    y_{\nu}
    \end{bmatrix}^T = 
    Q^{-1}
    \begin{bmatrix}
    -\nabla_u L \enskip 
    -\nabla_v L \enskip
    0 \enskip
    0
    \end{bmatrix}^T.
\end{align*}
%

Hence, the forward and backward passes with our module are respectively given by: 1) Using the expression in \eqref{eqn:newton}, solve for the logit equilibrium given $P$, and 2) Using $\nabla_u L$ and $\nabla_v L$, obtain $\nabla_P L$ using \eqref{eqn:gradL_P}. It is stressed that the module is sufficiently general to be included in any existing architecture where having a zero-sum game module is appropriate.
\subsubsection{A Note on Identifiability}
As mentioned in Section~\ref{sec:intro}, it is natural to ask if the games are identifiable -- that is, is there a unique $P$ which under the logit QRE, generates $u^*, v^*$? The answer is no, in general. Assuming $u^*, v^*$ are fixed, we can rewrite the KKT conditions in \eqref{eqn:KKT} as a system of linear equations in $P$. This system has $\mathcal{O}(nm)$ unknowns but only $\mathcal{O}(n+m)$ constraints. This implies that without a sufficiently compact parametrization, there will be infinitely many payoff matrices leading to identical equilibria. For example, one can add a constant to all entries in $P$ without changing the QRE. When under-constrained, one cannot expect to recover $P$. However, as we show below, there are also many settings where it \emph{is} possible to recover underlying parameters reliably.
\section{Learning Extensive Form Games (EFG)}
\subsection{Sequence Form Representation}
In practice, many games are more naturally and compactly represented in extensive form. Unfortunately, learning payoff matrices of their equivalent normal form representation is computationally unfeasible even for small games. For example, one-card poker has $2^{26}$ pure strategies per player. In order to facilitate learning of EFGs, we turn to the \textit{sequence form} representation~\cite{von1996efficient}, which is sufficiently rich to represent all strategic behaviors given perfect recall.

The sequence form replaces pure strategies by partial description of sequences specifying the player's moves over the game tree. Instead of probability vectors, we are interested in \textit{realization plans} $u \in \mathbb{R}^n, v\in \mathbb{R}^m$, each a vector of size equal to possible actions throughout the game (thus, $n$ and $m$ are equal to the size of the game tree's action nodes for each player). Realization plans represent probabilities of performing a sequence of actions, in isolation from chance and other player's moves. Mathematically, this is represented by the linear constraints
$Eu = e, Fv = f$.   Here, $E, F$ are matrices with entries in $\{-1,0,1\}$, while $e,f$ are vectors containing $\{0, 1\}$. Together, they specify `flow' constraints and implicitly encode parent-child relationships and information sets. These constraints may be seen as a generalization of the requirement that $u, v$ lie in probability simplexes. The likelihood of arriving at a given node of the game tree is the product of probabilities of each sequence, multiplied by the probabilities needed by the chance player.


\subsection{Dilated Entropy Regularization}
Denote $\mathcal{I}_u$ and $\mathcal{I}_v$ to be all information sets for the min and max player. For an information set $i \in \mathcal{I}_u \cup \mathcal{I}_v$, $\mathcal{A}_i$ denotes the possible actions at information set $i$, while $p_i$ is the action (from the same player) preceding $i$. Similarly, define $\rho_a$ to be the information set immediately preceding the action $a$ ,i.e. $i$ where $a \in \mathcal{A}_i$. As with the normal form representation, we solve the regularized min-max problem
\begin{align}
\min_{u} 
    \max_{v} \; & u^T P v + \sum_{i \in \mathcal{I}_u} \sum_{a \in \mathcal{A}_i} u_a \log \frac{u_a}{u_{p_i}}
    - \sum_{i \in \mathcal{I}_v} \sum_{a \in \mathcal{A}_i} v_a \log \frac{v_a}{v_{p_i}}.
\label{eqn:seq_regularization}
\end{align}
This form of regularization is known as dilated entropy \cite{kroer2017theoretical} or normalized entropy \cite{boyd2004convex}, and is known to be strictly convex/concave in $u$ and $v$ respectively. Observe that this formulation operates in $\mathcal{O}(m+n)$ dimensions. The number of sequences is bounded by the size of the game tree and is normally much smaller than the number of pure strategies.

One of our first primary results in this paper is the fact that this particular form of regularization, applied to the sequence form, recovers the QRE as applied to the equivalent \emph{reduced} normal form game (that is, the normal form representation of the extensive form, but with unattainable strategies omitted).

\begin{theorem}
The solution to \eqref{eqn:seq_regularization} is realization equivalent to the QRE of the game in reduced normal form. 
\label{thm:rnf}
\begin{proof} \emph{(Sketch)} Consider the max player in isolation and his game tree, represented by alternating actions and information states. Choose any action $a_0$ with parallel information sets, near the bottom of the tree (i.e. all child information sets are leaves). In sequence form, this action and its subtree may be coalesced into a set of strategies, corresponding to the Cartesian product $\prod_{\{i |p_i = a_0\}} \mathcal{A}_i$. This essentially converts a part of the sequence form into normal form. It can be shown that the solutions to \eqref{eqn:seq_regularization} before and after this replacement are realization equivalent. 
The proof follows by repeated bottom-up application of this operation, eventually collapsing the tree to its reduced normal form, all while maintaining realization equivalence. The full proof is in the Appendix.
\end{proof}
\end{theorem}
Theorem~\ref{thm:rnf} shows dilated entropy regularization leads to a well-accepted solution concept, even if it differs slightly from the more traditional definition of the QRE for extensive form games \cite{mckelvey1998quantal}.  A side consequence is that under certain regimes, the excessive gap technique \cite{kroer2017theoretical,hoda2010smoothing} used to quickly solve zero-sum EFGs converges to the NE specified by QRE in reduced normal form, as the rationality-parameter tends to $\infty$.

\subsection{Differentiable Learning in Sequence Form}
Here we derive a differentiable formulation of the sequence form QRE, mirroring our derivation for the normal form case, but admittedly with significantly more complex notation due to the more involved entropy term.
The KKT conditions of our optimization problem are,
\begin{eqnarray*}
    \lefteqn{(Pv)_a + 1 + \log(u_a) - \log(u_{p_i})} 
    \nonumber \\ &&{}- J_a 
    + \sum_{c \in \mathcal{C}_a} \mu_c - \mu_{i}
    = 0, \quad \forall i \in \mathcal{U}, a \in \mathcal{A}_i 
    \nonumber \\
    \lefteqn{(P^Tu)_{a'} - 1 - \log(v_{a'}) + \log(v_{p_{i'}})}
    \nonumber \\ &&{} + J_{a'} 
    + \sum_{c \in \mathcal{C}_{a'}} \nu_c - \nu_{i'}
    = 0, \quad \forall i' \in \mathcal{V}, a' \in \mathcal{A}_{i'}
    \nonumber \\
 	\lefteqn{Eu-e = 0, \quad
	Fv-f = 0, \quad u \ge 0, \quad v \ge 0}  \nonumber
\end{eqnarray*}
where $\mathcal{C}_a, \mathcal{C}_{a'}$ are sets of possible information sets immediately following $a$ or $a'$. $J_a,J_{a'}$ are their sizes, i.e. $|\mathcal{C}_a|, |\mathcal{C}_{a'}|$.

We write the terms on the left hand side as a vector $g(u, v, \mu, \nu)$. Taking derivatives again yields the updates for Newton's method.
\begin{align*}
    \begin{bmatrix}
    -\Xi(u) & P & E^T & 0 \\
    P^T & \Xi(v) & 0 & F^T \\
    E & 0 & 0 & 0 \\
    0 & F & 0 & 0
    \end{bmatrix}
    \begin{bmatrix}
    \Delta u \\ \Delta v \\ \Delta \mu \\ \Delta \nu
    \end{bmatrix}
    &= -g(u, v, \mu, \nu)\\
    \Xi(u)_{ab} = 
    \begin{cases}
    -\frac{1+J_a}{u_a},a=b\\
    \frac{1}{u_b},p_{\rho_a}=b\\
    \frac{1}{u_a},p_{\rho_b}=a
    \end{cases}
    \hspace{-0.5em}
    \Xi(v)_{a'b'} &= 
    \begin{cases}
    -\frac{1+J_{a'}}{v_{a'}}, a'=b'\\
    \frac{1}{v_{b'}}, p_{\rho_{a'}}=b'\\
    \frac{1}{v_{a'}}, p_{\rho_{b'}}=a'
    \end{cases}
\end{align*}
The updates are done in exactly the same manner as \eqref{eqn:gradL_P} 
\begin{align}
\label{eqn:seq_gradL_P}
    \nabla_P L = y_u v^T + u y_v^T,
\end{align}
where
\begin{align*}
    \begin{bmatrix}
    y_u \\
    y_v \\
    y_{\mu} \\
    y_{\nu}
    \end{bmatrix} &= 
    \begin{bmatrix}
    -\Xi(u) & P & E^T & 0 \\
    P^T & \Xi(v) & 0 & F^T \\
    E & 0 & 0 & 0 \\
    0 & F & 0 & 0
    \end{bmatrix}^{-1}
    \begin{bmatrix}
    -\nabla_u L \\ 
    -\nabla_v L \\ 
    0 \\ 
    0
    \end{bmatrix}.
\end{align*}
\subsubsection{Implementation Notes}
We can use this differentiable game solver within an automatic differentiation framework to easily obtain gradients of virtually any loss with respect to any of the game parameters.  In particular, we used the PyTorch automatic differentiation library \cite{paszke2017automatic}, and will release the full code for our solver as open source along with the release of this paper.

\section{Experiments}
We empirically demonstrate our module's novel aspects -- learning \textit{extensive form games} in the presence of \textit{side information}, with \textit{partial observations}. In the first experiment, we learn a non-symmetric variant of rock, paper, scissors with side information. We illustrate the learning of extensive form games with one-card poker, and learning with partial information with a security resource allocation game. In all cases, we minimize the log-loss of the observed sequence - that is, maximizing the likelihood of realizing observed \textit{sequence} from the player, assuming he acts in accordance to the QRE.

Due to space constraints, hyper-parameters and details of train/test environment are deferred to the Appendix. Generally, our module works well with a medium or large batch size (e.g. 128), RMSProp \cite{Tieleman2012} or Adam \cite{kingma2014adam} optimizers with learning rates between $[0.0001 ,0.01]$.
\subsection{Rock, Paper, Scissors}

Rock Paper Scissors (RPS) is among the most well-studied 2-player zero-sum game. It is well known that playing uniformly is an NE and QRE for RPS. In this experiment, we consider the following variant (Figure~\ref{fig:modified_RSP}), which breaks symmetry between the 3 actions. Notice that the traditional RPS is recovered when $b_1,b_2,b_3$ are all 1.


\begin{figure}
    \centering
    \begin{tabular}{| c | c | c | c |}
    \hline
      & R & P & S \\ \hline
    R & $0$ & $-b_1$ & $b_2$ \\ \hline
    P & $b_1$ & $0$ & $-b_3$\\ \hline
    S & $-b_2$ & $b_3$ & $0$ \\
    \hline
  \end{tabular}
    \caption{Payoff matrix of modified Rock-Paper-Scissors. Values shown are for the row player.}
    \label{fig:modified_RSP}
\end{figure}

\begin{figure}
\centering
    \minipage{0.25\textwidth}
    \includegraphics[scale=0.26]{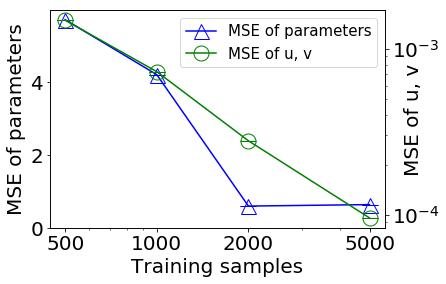}
    \endminipage
    \minipage{0.25\textwidth}
    \includegraphics[scale=0.26]{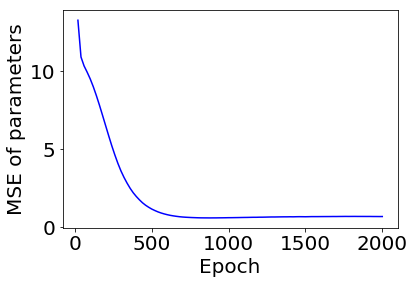}
    \endminipage
    \caption{Left: MSE of parameters and predicted mixed strategies. Right: Convergence over $2000$ epochs, using a dataset of size $2000$.}
   \label{fig:exp_RPS}
\end{figure}

We assume that each of the $b$'s is a linear function of some features $x \in \mathbb{R}^2$, i.e., $b_y=x^Tw_y, y\in\{1,2, 3\}$, where $w_y$ are to be learned. 
Features in the dataset and ground-truth weights are drawn uniformly from $[0, 1]$, and $[0, 10]$ respectively.
Experiments were evaluated with a fixed test set of size $2000$. 
The results presented in Figure~\ref{fig:exp_RPS} illustrate 2 key points. The first plot shows that dramatically improves with larger datasets, both in terms of parameters learned and predicted strategies. The second plot shows that with a reasonably sized dataset, convergence is stable and is fairly quick. However, it was observed that when the dataset is small (e.g. 200), the model may diverge from the ground truth. 
\subsection{One-Card Poker}

We consider a simple poker game where players are dealt a single card, with ante/bets of $10$, and two stages of betting for the first player. Specifically, suppose $n$ cards labeled $1$ through $n$ are dealt uniformly. Both players begin with an ante of $10$. Player 1 decides whether to bet an additional $10$, followed by Player 2 (who folds if he does not call). Lastly, Player 1 may choose to bet if Player 2 raises. Both players are obliged to reveal their card at the end of each game. 

While relatively simple, the game contains the key elements in extensive-form games and the strategic concepts in poker such as slow playing (e.g. not betting even if Player 1 holds the high card) and bluffing (e.g. betting even if the player holds a low card). Despite its simple structure, the game with $n$ cards has $2^{2n}$ normal form pure strategies for each player. This exponential explosion of strategies is not improved by using the reduced normal form. However, in sequence form, we only need to work with realization plans of size $4n$. 
\subsubsection{Experimental Setup} We assume that the deck is stacked non-uniformly. Our goal is to learn this distribution of cards after observing many rounds of play. Let $d \in \mathbb{R}^n, d \ge 0, \sum_i d_i = 1$ be the \textit{weights} of cards. 
The probability that the players are dealt cards $(i,j)$ ($i\neq j$) is $d_i \times \frac{d_j}{1-d_i}$. Note that this distribution is asymmetric between players. Matrices $P, E, F$ for the case $n=4$ are presented in the Appendix.

\textit{Remark.} While counting cards seems to be a straightforward way to learn the card distribution when $d$ does not change over time, our method is suited to learn the player's \textit{perceived} or \textit{believed} distribution of cards, which may be different from the distribution of cards dealt. This may even be a function of contextual features such as demographics of players.

\begin{figure}
\centering
    \minipage{0.25\textwidth}
    \includegraphics[scale=0.27]{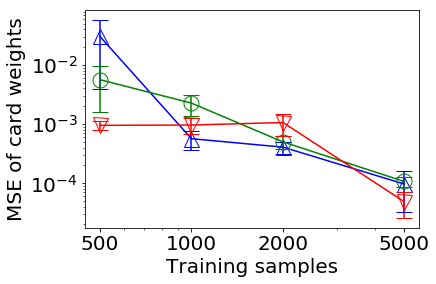}
    \endminipage
    \minipage{0.25\textwidth}
    \includegraphics[scale=0.27]
{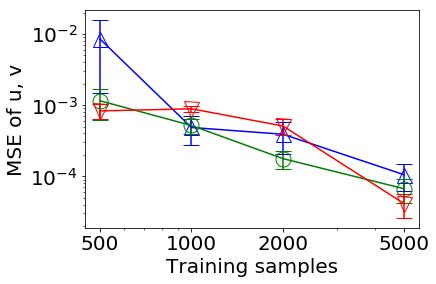}
    \endminipage
    \caption{MSE of card weights (left) and realization plans (right).}
   \label{fig:poker}
\end{figure}

A total of three experiments were run with $n=4$. For each experiment, $d \sim \text{Dir}(1,1,1,1)$. Each experiment comprises 5 runs of training, with same weights but different training sets. Training was for $2500$ epochs, which was observed to be after convergence. The mean squared error of learned parameters are averaged over all runs and are presented in Figure~\ref{fig:poker}.

\subsection{Security Resource Allocation Game}

In this set of experiments, we demonstrate the ability to learn from \emph{incomplete observations} in a setting that abstracts attacks in cybersecurity domain. The defender possesses $k$ indistinguishable and indivisible defensive resources, e.g., cyber analysts, which he splits among $n$ targets, $\{T_1, ... ,T_n\}$. In an attacking attempt, the attacker (row player) chooses one target. In the event an attack on $T_i$ succeeds, the attacker obtains a reward of $R_i$ (and the defender $-R_i$), otherwise, the payoffs to both parties are $0$. Each defensive resource independently prevents an intrusion with probability $0.5$. For example, if there are two defenders guarding $T_1$, the chance of a successful attack on $T_1$ is $\frac{1}{2^2}$. This creates a scenario where the marginal benefit of each defensive resources decreases, thus requiring the defender to strike a balance. The matrix of expected payoffs when $n=2, k=3$ is shown in Figure~\ref{fig:teamselection}.
\begin{figure}
    \centering
    \begin{tabular}{| c | c | c | c | c|}
    \hline
    $\{\# D_1, \# D_2 \}$ & \{0, 3\} & \{1, 2\} & \{2, 1\} & \{3, 0\}\\ \hline
    $T_1$ & $-R_1$ & $-\frac{1}{2}R_1$ & $-\frac{1}{4}R_1$ & $-\frac{1}{8}R_1$\\ \hline
    $T_2$ & $-\frac{1}{8}R_2$ & $-\frac{1}{4}R_2$ & $-\frac{1}{2}R_2$ & $-R_2$\\ 
    \hline
  \end{tabular}
    \caption{Security resource allocation game. $n=2, k=3$.
    }
    \label{fig:teamselection}
\end{figure}

In addition to considering the case where the attacker launches a single attack, we also consider a multi-stage game where the attacker can launch $t$ attacks, one in each stage while the defender chooses his allocation of resources in stage $1$ and cannot change it in later stages. On the other hand, the attacker has the \textit{option} of changing his target between stages. This describes a setting where analysts are deployed to specific network assets on a daily basis, while attackers are sufficiently nimble to make multiple attacks in a single day. To understand why the attacker may change target, consider that target $T_i$ is attacked in stage $1$ and the attack is unsuccessful. It may be inferred that it is more likely that $T_i$ is better guarded, prompting the attacker to switch targets.

\begin{figure}
\centering
    \minipage{0.25\textwidth}
    \includegraphics[scale=0.27]{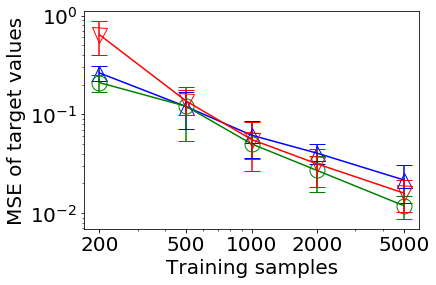}
    \endminipage
    \minipage{0.25\textwidth}
    \includegraphics[scale=0.27]{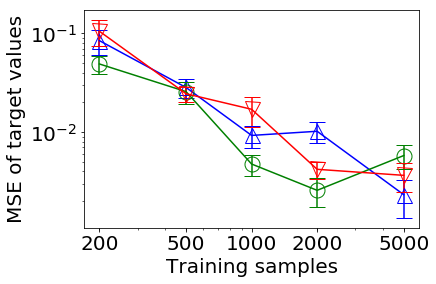}
    \endminipage\\
    \caption{MSE of target values. Left: t=1, Right: t=2}
   \label{fig:teamsel_results}
\end{figure}

Three experiments are run with $n=2, k=5$ for games with single attack and double attack, i.e, $t=1$ and $t=2$. Crucially, in this set of experiments, we learn $R_i$ only based on observations of the defender's actions. 
This setting yields a $10 \times 6$ sequence form payoff matrix. For each experiment, $R_1$ and $R_2$ are drawn uniformly in $[0,2]$. Each experiment is run $10$ times for at least $2000$ epochs per run. 
The mean and standard error over each run is presented in Figure~\ref{fig:teamsel_results}. The results show that our algorithm can still recover the game setting by only observing defender's actions.
\subsection{Discussion} As expected, the quality of learned parameters improves as the number of data points increases. Notable exceptions occur in (i) the green plot for the security game when $t=2$ and (ii) RPS, when comparing between training sizes of $2000$ and $5000$. These outliers are no longer observed when comparing MSE of $u, v$. For example, Figure~\ref{fig:exp_RPS} shows that predicted strategies improve significantly when going from $2000$ to $5000$ samples, showing that despite not converging to better parameters, the network still demonstrates a marked improvement in predicting player strategies. 
\section{Conclusion}
In this paper, we present a fully differentiable module capable of learning payoff and other parameters in zero-sum games, given side information and partial observability. The proposed module's unique capabilities are demonstrated over a broad range of problems. Future work entails faster solvers by exploiting structure in the KKT matrix and extensions to learning general-sum games.


\appendix

\bibliographystyle{named}
\bibliography{ijcai18}
\newpage
\section{Proofs and derivations}

\subsection{Equivalence to reduced normal form}
In this section we present the proof of one of the main technical results of this paper, that the dilated entropy regularization of a extensive form game is equivalent to the standard entropy-regularized QRE of the equivalent reduced normal-form game. 

Consider the max player in isolation. His game in sequence form may be represented by a tree with alternating actions and information sets. Consider the game shown in Figure~\ref{fig:reduced_equals_reg_simple}. Red nodes are vertices and black nodes are information sets. White squares represent sibling actions of $A$, which may in turn, contain other parts of the game tree. The information states associated with $I_1, I_2$ are parallel information states, i.e. both of which may be reached from $A$ with non-zero probability (assuming suitable opponent/chance actions). 

We show that the optimal solution using dilated entropy regularization on the game in Figure~\ref{fig:reduced_equals_reg_simple} is realization equivalent to the optimum of the reduced game in Figure~\ref{fig:reduced_equals_reg_simple_after}. Essentially, this operation converts \textit{part} of the sequences to the normal form, by replacing parts of the tree with all possibly contingencies. Repeated application of this operation eventually translates the sequence-form representation to the reduced normal form. This is performed on the tree in a bottom-up manner, all while maintaining realization equivalence.

\tikzset{
  treenode/.style = {align=center, inner sep=0pt, text centered,
    font=\sffamily},
  arn_n/.style = {treenode, circle, black, font=\sffamily\bfseries, draw=none,
    fill=white, text width=1.5em},
  arn_r/.style = {treenode, circle, red, draw=none, 
    text width=1.5em, very thick},
  arn_x/.style = {treenode, rectangle, draw=black,
    minimum width=0.5em, minimum height=0.5em}
}
\usetikzlibrary{arrows}
\begin{figure}[h]
\begin{tikzpicture}[->,>=stealth',level/.style={sibling distance = 4.0cm/#1,
  level distance = 2.0cm}] 
\node [arn_n] {I}
    child{ node [arn_r] {A} 
            child{ node [arn_n] {I1} 
            	child{ node [arn_r] {11}}
							child{ node [arn_r] {12}}
            }
            child{ node [arn_n] {I2}
							child{ node [arn_r] {21}}
							child{ node [arn_r] {22}}
            }
    }
    child{ node [arn_x] {} 
    }
; 
\end{tikzpicture}
\caption{Original sequence form.}
\label{fig:reduced_equals_reg_simple}
\end{figure}
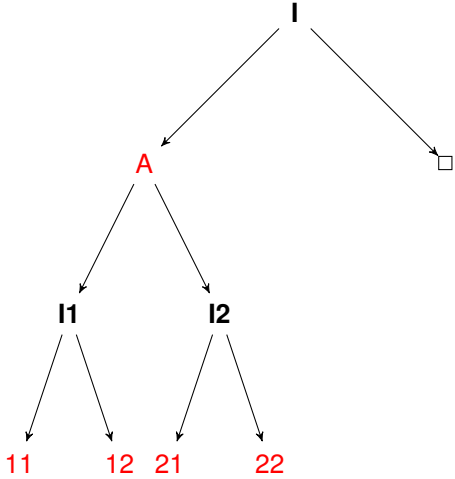
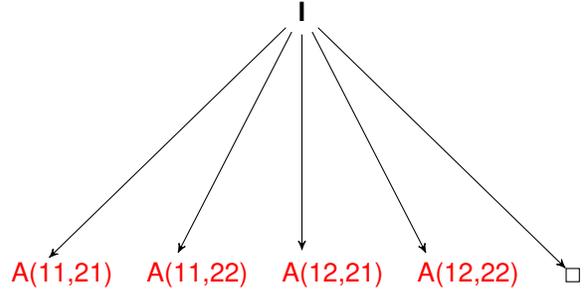
\begin{figure}[h]
\begin{tikzpicture}[->,>=stealth',level/.style={sibling distance = 1.8cm/#1,
  level distance = 3.5cm}] 
\node [arn_n] {I}
    child{ node [arn_r] {A(11,21)} 
    }
    child{ node [arn_r] {A(11,22)} 
    }
    child{ node [arn_r] {A(12,21)} 
    }
    child{ node [arn_r] {A(12,22)} 
    }
    child{ node [arn_x] {} 
    }
; 
\end{tikzpicture}
\caption{After a single iteration.}
\label{fig:reduced_equals_reg_simple_after}
\end{figure}

\subsubsection{Part 1}
Let $A$ be the an action such that all immediate child \textit{actions} are leaves (e.g. Figure~\ref{fig:reduced_equals_reg_simple}) and $I$ be its parent. We \textit{condition} on having taken the action $A$, that is, assume that the player reaches $A$ with probability $1$ (barring chance or the other player's actions). We will first show that after conditioning, the resultant reduced normal form is equivalent to its sequence form. 

\textbf{Notation.} 
Recall $\mathcal{C}_A = \{ I_k \}$ denotes the children \textit{information sets} of $A$. The action set at $I_k$ is denoted by $\mathcal{A}_k = \{A_{kj}\} = \{ a \in \mathcal{A}_v | \rho_a = I_k \}$. For the purposes of this derivation, the opponent's strategy is fixed, and his strategy profile is combined to give $P^Tu = x$. 

In normal form, all contingencies are accounted for, and the set of pure strategies are given by the Cartesian product of action sets $\hat{\mathcal{A}}_A = \prod_{k} \mathcal{A}_k$. We denote the normal form mixed strategy profile, payoff matrix, and normal form payoff vector as $\hat{v}$, $\hat{P}$ and $\hat{x}=\hat{P}^T\hat{u}$. Note that the sizes of these vectors/matrices are not equivalent to their counterparts in sequence form. For convenience, we define $f(\hat{v})$ to be a function mapping $\hat{v}$ to $v$ by performing the appropriate marginalization. Let $g(v, j)$ be the distribution of actions extracted from $v$ supposing an information state of $I_j$. For simplicity let $H$ now define the Shannon entropy, $H(y) = -\sum y_i\log (y_i)$.

\textbf{Derivation.}
The resultant QRE in the reduced normal form is,
\begin{align*}
    & \max_{\hat{v}} \hat{v}^T\hat{x} + H(\hat{v}) \\
    =& \max_{\hat{v}} f(\hat{v})^T x + H(\hat{v}) \\
    =& \max_{\hat{v}} f(\hat{v})^Tx + H \left( \bigotimes_{j \in \mathcal{C}_A} g(f(\hat{v}), j) \right) \\
    =& \max_{\hat{v}} f(\hat{v})^Tx + \sum_{j \in \mathcal{C}_A} H(g(f(\hat{v}), j)) \\
    =& \max_{v} v^Tx + \sum_{j \in \mathcal{C}_A} H(g(v, j)).
\end{align*}
The first line is by definition. The second line holds from the fact that joint entropy is maximized by independent random variables (induced by the marginals $g(f(\hat{v}), j)$). That is, if this independence relationship does not hold, then we could construct a strictly better candidate solution. The third line follows from the fact that the joint entropy of mutually independent random variables is equal to the sum of their entropies. The last line is true since every sequence form strategy is induced by at least one normal form strategy (i.e. the image of $f$ is equal to the domain of $v$) and vice versa. 

\subsubsection{Part 2}
In Part 1, we showed realization equivalence when the white square is non-existent, i.e. there are no other children of $I$. In this section, we relax that assumption. For this part of the proof, $I$ \textit{may} possibly not be the root. 

Consider the \textit{full} sequence form formulation, starting from our objective function. We split $v$ (the full realization plan in sequence form) into two portions, $\tilde{v}$ for actions which are children of $A$ (i.e. all $a$ where $p_{\rho_a} = A$), and $\acute{v}$, for the rest of the actions (at other portions of the tree). We split $x$ into $\tilde{x}$ and $\acute{x}$ similarly. The objective function may be written as two nested optimization problems,

\begin{align*}
&\max_v v^T x - \sum_{i \in \mathcal{I}_v} \sum_{a \in \mathcal{A}_i} v_a \log \frac{v_a}{v_{p_i}}\\
=& \max_{\acute{v}} \acute{v}^T \acute{x} - \sum_{p_i \neq A} \sum_{a \in \mathcal{A}_i} v_a \log \frac{v_a}{v_{p_i}} \\
& \quad + \tilde{v}^T \tilde{x} -  \sum_{p_i = A} \sum_{a \in \mathcal{A}_i} v_a \log \frac{v_a}{v_{p_i}}  
\end{align*}

Observe that the flow constraints dictate that for any information set under $A$, the corresponding children actions in $\tilde{v}$ sums to $v_A$. Hence, we may write the optimization problem in terms of normalized (conditional) probabilities, $\bar{v}$, which \textit{do} indeed sum to 1.
\begin{align*}
& \max_{\acute{v}} \Bigg( \acute{v}^T \acute{x} - \sum_{p_i \neq A} \sum_{a \in \mathcal{A}_i} \acute{v}_a \log \frac{\acute{v}_a}{\acute{v}_{p_i}} \\
& \quad + \acute{v}_A \max_{\bar{v}} \left( \bar{v}^T \bar{x} -  \sum_{p_i = A} \sum_{a \in \mathcal{A}_i} \bar{v}_a \log \bar{v}_a  \right) \Bigg)
\intertext{After normalization, we may plug in the result in Part 1 into the inner maximization term, replacing it by its normal form involving $\hat{v}$ and $\hat{x}$.}
& \max_{\acute{v}} \Bigg( \acute{v}^T \acute{x} - \sum_{p_i \neq A} \sum_{a \in \mathcal{A}_i} \acute{v}_a \log \frac{\acute{v}_a}{\acute{v}_{p_i}} \\
& \quad + \acute{v}_A \max_{\hat{v}} \left( \hat{v}^T \hat{x} + H(\hat{v}) \right) \Bigg) \\
=&\max_{\acute{v}} \Bigg( \acute{v}^T \acute{x} - \sum_{\substack{p_i \neq A \\ i \neq \rho_A}} \sum_{a \in \mathcal{A}_i} \acute{v}_a \log \frac{\acute{v}_a}{\acute{v}_{p_i}} -\sum_{\substack{a \in \mathcal{A}_{\rho_A} \\ a \neq A}} \acute{v}_a \log \frac{\acute{v}_a}{\acute{v}_{p_{\rho_A}}}\\
& \quad - \acute{v}_A \log \frac{\acute{v}_A}{\acute{v}_{p_{\rho_A}}} + \acute{v}_A \max_{\hat{v}} \left( \hat{v}^T \hat{x} + H(\hat{v}) \right) \Bigg) \\
=&\max_{\acute{v}} \Bigg( \acute{v}^T \acute{x} - \sum_{\substack{p_i \neq A \\ i \neq \rho_A}} \sum_{a \in \mathcal{A}_i} \acute{v}_a \log \frac{\acute{v}_a}{\acute{v}_{p_i}} -\sum_{\substack{a \in \mathcal{A}_{\rho_A} \\ a \neq A}} \acute{v}_a \log \frac{\acute{v}_a}{\acute{v}_{p_{\rho_A}}}\\
& \quad - \acute{v}_A \log \frac{\acute{v}_A}{\acute{v}_{p_{\rho_A}}} + \max_{\sum \hat{v} = \acute{v}_A} \left( \hat{v}^T \hat{x} - \sum_{\hat{v}} \hat{v} \log \frac{\hat{v}}{\acute{v}_{A}} \right) \Bigg)
\end{align*}

The first line follows from the result in Part 1. The second line splits the entropy regularization terms from $I$ into the terms involving $A$ and those which do not. The last line follows from `reintroducing' the flow constraints into the inner maximization. Observe that the $\acute{v}_A \log \acute{v}_A$ terms cancel out, allowing for the following simplifications,

\begin{align*}
&\max_{\acute{v}} \Bigg( \acute{v}^T \acute{x} - \sum_{\substack{p_i \neq A \\ i \neq \rho_A}} \sum_{a \in \mathcal{A}_i} \acute{v}_a \log \frac{\acute{v}_a}{\acute{v}_{p_i}} -\sum_{\substack{a \in \mathcal{A}_{\rho_A} \\ a \neq A}} \acute{v}_a \log \frac{\acute{v}_a}{\acute{v}_{p_{\rho_A}}}\\
& \quad - \acute{v}_A \log \frac{1}{\acute{v}_{p_{\rho_A}}} + \max_{\sum \hat{v} = \acute{v}_A} \left( \hat{v}^T \hat{x} - \sum_{\hat{v}} \hat{v} \log \hat{v}\right) \Bigg) \\
=&\max_{\acute{v}} \Bigg( \acute{v}^T \acute{x} - \sum_{\substack{p_i \neq A \\ i \neq \rho_A}} \sum_{a \in \mathcal{A}_i} \acute{v}_a \log \frac{\acute{v}_a}{\acute{v}_{p_i}} -\sum_{\substack{a \in \mathcal{A}_{\rho_A} \\ a \neq A}} \acute{v}_a \log \frac{\acute{v}_a}{\acute{v}_{p_{\rho_A}}}\\
& \quad \max_{\sum \hat{v} = \acute{v}_A} \left( - \sum_{\hat{v}} \hat{v} \log \frac{1}{\acute{v}_{p_{\rho_A}}} +  \hat{v}^T \hat{x} - \sum_{\hat{v}} \hat{v} \log \hat{v}\right) \Bigg) \\
=&\max_{\acute{v}} \Bigg( \acute{v}^T \acute{x} - \sum_{\substack{p_i \neq A \\ i \neq \rho_A}} \sum_{a \in \mathcal{A}_i} \acute{v}_a \log \frac{\acute{v}_a}{\acute{v}_{p_i}} -\sum_{\substack{a \in \mathcal{A}_{\rho_A} \\ a \neq A}} \acute{v}_a \log \frac{\acute{v}_a}{\acute{v}_{p_{\rho_A}}}\\
& \quad \max_{\sum \hat{v} = \acute{v}_A} \left( - \sum_{\hat{v}} \hat{v} \log \frac{\hat{v}}{\acute{v}_{p_{\rho_A}}} +  \hat{v}^T \hat{x} \right) \Bigg).
\end{align*}

Observe that $\acute{x}_A$ is zero, since it must be a non-leaf node (all rewards are deferred to leaves). This allows us to recombine $\acute{v}$ and $\hat{v}$ into a long vector in sequence form, which we denote $\grave{v}$ -- which is the sequence form of the condensed version (e.g. Figure~\ref{fig:reduced_equals_reg_simple_after}. The flow constraints in the inner maximization is `compatible' with the form required by the outer maximization.

\begin{align*}
\max_{\grave{v}} \grave{v}^T \grave{x} - \sum_{i \in \grave{\mathcal{I}}_{v}} \sum_{a \in \grave{\mathcal{A}}_i} \grave{v}_a \log \frac{\grave{v}_a}{\grave{v}_{p_i}}
\end{align*}
where the $\grave{\mathcal{A}}$ and $\grave{\mathcal{I}_v}$ are the action sets in the condensed game (i.e all the information sets originally children of $A$ are gone, and the actions in the normal form are now actions in $\rho_A$). $\grave{x}$ refers to the reward vector obtained by merging $\hat{x}, \acute{x}$ together, and removing the term associated with $\acute{x}_A$ (which was $0$ to begin with). 

The final part of the proof follows by repeatedly applying the transform described. Note that it is always possible to find a suitable $I$ and $A$ for this process, if not, then we have already arrived at the reduced normal form. The number of non-leaf actions must drop monotonically, hence the process will terminate at some point. Lastly, observe that our assumption that non-leaf actions have $0$ payoffs is remains true after every iteration.

\section{Experimental setup}
\subsection{E, F, P matrices for one-card poker}
Here we provide a complete example of how to encode and parameterize one-card poker in sequence form, as required by our approach. We begin with the linear constraints in sequence form. Recall $E$ and $F$ matrices define the structure of the game state transitions, $Eu-e=0$, $Fv-f=0$. For a game with $4$ cards, there are $8$ information sets and $16$ actions per player. The constraint matrices are given by
$$
E = \begin{bmatrix} I & I & 0 & 0 \\ -I & 0 & I & I 
\end{bmatrix}
$$
$$F = 
\begin{bmatrix}
I & 0 & I & 0\\
0 & I & 0 & I
\end{bmatrix}
$$
$$ e = [1 1 1 1 1 1 1 1 0 0 0 0 0 0 0 0]^T $$
$$ f = [1 1 1 1 1 1 1 1 1 1 1 1 1 1 1 1]^T $$
where each $I$ has a size of $4$. 

We now turn to parameterization of the payoff matrix in terms of the card distribution $d$. Suppose for the time being that the deck is uniformly stacked, i.e. $d=(0.25, 0.25, 0.25, 0.25)$. Define the \textit{showdown} matrix
$$ S = 
\frac{1}{12} \begin{bmatrix}
0 & 1 & 1 & 1 \\
-1 & 0 & 1 & 1 \\
-1 & -1 & 0 & 1 \\
-1 & -1 & -1 & -1
\end{bmatrix}
$$
and the \textit{forfeit} matrix.
$$ X = 
\frac{1}{12}\begin{bmatrix}
0 & 1 & 1 & 1 \\
1 & 0 & 1 & 1 \\
1 & 1 & 0 & 1 \\
1 & 1 & 1 & 0
\end{bmatrix}
$$.
Then 
$$ P = 
\begin{bmatrix}
S & 0 & 0 & 0 \\
0 & 0 & -X & 2S \\
0 & X & 0 & 0 \\
0 & 2S & 0 & 0
\end{bmatrix}
$$
where by our convention, the row player is the minimizing player. Each block is of size $4 \times 4$, and gives possible outcomes for each of the $4^2$ possible ways of dealing cards. The 4 blocks for the row player correspond to the actions `do not raise on first move', `raise on first move', `fold on second move', `raise on second move'. For the column player, the $16$ actions are `fold after first player did not raise', `raise after first player did not raise', `fold after first player raised', `raise after first player raised'.  

When $d$ is not uniform, define the $4 \times 4$ distribution matrix 
$$
D_{ij} = \frac{d_i d_j}{1-d_i}.  
$$ 
The payoff matrix is then 
$$
\begin{bmatrix}
D \circ S & 0 & 0 & 0 \\
0 & 0 & -1 \circ D & 2 D \circ S \\
0 & 1 \circ D & 0 & 0 \\
0 & 2 D \circ S & 0 & 0
\end{bmatrix}
$$
i.e. every $4 \times 4$ block is pointwise weighted by the chance of being dealt the relevant cards. It may be seen that the forfeit matrix $X$ is really just the all-ones matrix multiplied pointwise by $D$.
\subsection{Experimental setup}
All of the experiments are run using CPU cycles. 
\subsubsection{Rock, paper, scissors}
Experiments were run on a 3.1 GHz Intel Core i5 with 16 GB of RAM. The learning rate is $0.0005$, with a batch size of $128$. We utilized the Adam optimizer. The maximum number of epochs before termination is $10000$. Parameters were initialized to the $0$-matrix for each experiment.
\subsubsection{One card poker}
Experiments are run on a 4.2GHz Intel Core i7 with 128GB of RAM . The learning rate is $0.002$, batch size of $128$, using the RMSProp optimizer. Weights are initialized to be uniform, i.e. $(0.25, 0.25, 0.25, 0.25)$. In order to ensure that $d$ is valid probability distribution, the features are passed through softmax layer which then outputs $d$. The maximum number of epochs is $2500$, although convergence occurs significantly faster. Since there is no context, experiments may be run much faster by computing the forward pass just once for each minibatch. Similarly, the inverse matrix required in the backward pass may be cached and reused between each member in the same minibatch. 
\subsubsection{Security resource allocation game}
The experiments were run on an Amazon c4.2xlarge EC2 instance. The learning rate is $0.002$ using the RMSProp optimizer (all other hyperparameters are left as the defaults in Pytorch). Each run was $2000$ epochs. The weights are passed through $f(x)=(\text{tanh}(x) + 1)$ to clip rewards to between $[0, 2]$ for the payoff matrix.
\end{document}